\documentclass[conference]{IEEEtran}
\IEEEoverridecommandlockouts
\usepackage[caption=false]{subfig}
\usepackage{cite}
\usepackage{comment}
\usepackage{xcolor}
\usepackage{balance}
\usepackage{lipsum}     
\usepackage{ntheorem}   
\usepackage{mdframed}   

\newtheorem{Claim}{Claim}

\usepackage{kantlipsum}
\usepackage[breakable, theorems, skins]{tcolorbox}
\tcbset{enhanced}

\DeclareRobustCommand{\mybox}[2][gray!20]{%
\begin{tcolorbox}[   
        breakable,
        left=0pt,
        right=0pt,
        top=0pt,
        bottom=0pt,
        colback=#1,
        colframe=#1,
        width=9cm, 
        enlarge left by=0mm,
        boxsep=5pt,
        arc=0pt,outer arc=0pt,
        ]
        #2
\end{tcolorbox}
}
\theoremstyle{break}
\theoremheaderfont{\bfseries}
\newmdtheoremenv[%
linecolor=gray,leftmargin=60,%
rightmargin=40,
backgroundcolor=gray!40,%
innertopmargin=0pt,%
ntheorem]{myprop}{Proposition}[section]
\usepackage{amsmath,amssymb,amsfonts}

\usepackage{graphicx}
\usepackage{textcomp}
\newcommand{\aggregate}[2]{\underset{#2}{\operatornamewithlimits{#1\ }}}
\usepackage{algorithm}
\usepackage{algcompatible}
\usepackage{algpseudocode}

\usepackage{subfig}
 
\usepackage{xcolor}
\def\BibTeX{{\rm B\kern-.05em{\sc i\kern-.025em b}\kern-.08em
    T\kern-.1667em\lower.7ex\hbox{E}\kern-.125emX}}
\begin{document}

\title{Information-theoretic Feature Selection via Tensor Decomposition and Submodularity}

\author{\IEEEauthorblockN{Magda Amiridi}
\IEEEauthorblockA{\textit{Department of ECE } \\  
\textit{University of Virginia}\\
ma7bx@virginia.edu}
\and
\IEEEauthorblockN{Nikos Kargas}
\IEEEauthorblockA{\textit{Department of ECE } \\  
\textit{University of Minnesota}\\
karga005@umn.edu}
\and
\IEEEauthorblockN{Nicholas D. Sidiropoulos}
\IEEEauthorblockA{\textit{Department of ECE } \\  
\textit{University of Virginia}\\
nikos@virginia.edu}
}
\maketitle
\begin{abstract}
Feature selection by maximizing high-order mutual information between the selected feature vector and a target variable is the gold standard in terms of selecting the best subset of relevant features that maximizes the performance of prediction models. However, such an approach typically requires knowledge of the multivariate  probability distribution of all features and the target, and involves a challenging combinatorial optimization problem. Recent work has shown that any joint Probability Mass Function (PMF) can be represented as a naive Bayes model, via Canonical Polyadic (tensor rank) Decomposition. In this paper, we introduce a low-rank tensor model of the joint PMF of all variables and indirect targeting as a way of mitigating complexity and maximizing the classification performance for a given number of features.  Through low-rank modeling of the joint PMF, it is possible to circumvent the curse of dimensionality by learning ‘principal components’ of the joint distribution. By indirectly aiming to predict the latent variable of the naive Bayes model instead of the original target variable, it is possible to formulate the feature selection problem as maximization of a monotone submodular function subject to a cardinality constraint – which can be tackled using a greedy algorithm that comes with performance guarantees. Numerical experiments with several standard datasets suggest that the proposed approach compares favorably to the state-of-art for this important problem.
\end{abstract}
\section{Introduction}
Real-world data often exhibit complicated manifold structure in very high dimensional spaces, making conventional machine learning tools insufficient for data analysis. Knowledge discovery in a high-dimensional space with limited training examples is a difficult task that entails high computational cost in both the training and the run-time stage, large variance of the predictions due to overfitting the training samples, and poor generalization. Although adding more input variables may provide additional information, an assumption supported by the data processing inequality~\cite{CoTho1991}, after a certain point the performance of classification will typically degrade as the number of features continues to increase. In practice, not all features are equally important and discriminative, as many of the dimensions carry little or redundant information. Analyzing high-dimensional data therefore raises the fundamental problem of reducing dimensionality by discovering compact representations that do not incur significant loss in prediction accuracy for the ultimate task at hand. 
Feature selection methods try to find a lower-dimensional representation of data by removing redundant, irrelevant, or unimportant features. Feature selection maintains the physical meaning and dependencies between the selected features, resulting in predictive models with better interpretability~\cite{TaAleLiu2014,LiCheWa2018}. Feature selection aids the learning task since it aims to identify a feature subset of minimal size that is collectively optimally predictive with respect to the variable of interest, while also speeding up the computation time.

Projecting data onto a lower dimensional space facilitates, among other tasks, exploratory data analysis and visualization, clustering, and compression of high-dimensional data. Feature selection is particularly important and challenging in biomedical data mining, where the data is characterized by relatively few training instances and a high-dimensional feature space, leading to degradation of classifier performance as noisy/uninformative features prohibit us from mining potentially useful knowledge~\cite{RiHaBa2001}.
In personalized marketing, feature selection is used for sentiment analysis of customer reviews as it aims to identify indicators in the document to infer the polar category, either positive or negative sentiment, so that products are targeted to customers where the probability of positive sentiment is higher~\cite{DuSo2012}. 
Feature selection can be used in stock market price index prediction to reduce the cost of training time and to improve prediction accuracy~\cite{LungYi2009}. Feature selection has also been applied for improving text data clustering and classification~\cite{AgZha2012}.


To evaluate any possible subset, feature selection methods require a feature quality measure. Most prior information-theoretic methods for feature selection use a lower order approximation of the Mutual Information (MI). We consider using the high-order Shannon-entropy-based MI as the evaluation criterion due to the fact that it can capture any kind of relationship, both linear and nonlinear, between multiple random variables. Computing MI requires the estimation of a high-dimensional probability distribution. Direct estimation of the joint distribution for high-dimensional data is impossible, due to the curse of dimensionality. We thus need a `universal' model that can capture the `principal components' of this high-dimensional joint distribution in a parsimonious way. It has recently been shown that low-rank approximation of the joint probability tensor addresses this need~\cite{KaSiFu2018}. Rank-$F$ approximation represents the joint distribution as 
a latent variable model with just one hidden variable having $F$ possible states. For large enough but finite $F$ the latter model is universal -- it can represent {\em any} joint distribution of categorical variables. In this paper, we propose a novel dimensionality reduction framework that incorporates a low-rank model of the joint distribution, which affords disciplined subset selection through maximization of a monotone submodular function. The latter optimization is amenable to greedy solution with performance guarantees.
\section{Preliminaries}
\subsection{Canonical Polyadic Decomposition}
\begin{figure}[H]
\centering
\includegraphics[width = 0.8\columnwidth]{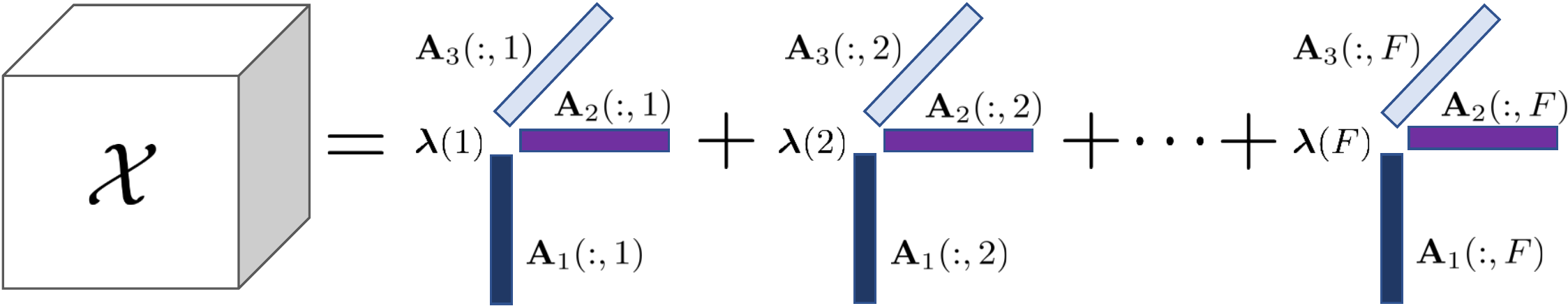}
\caption{CPD model of a $3$-way tensor with $F$ components.}
\label{CPD}
\end{figure}

Tensors provide a natural representation for massive multidimensional data. An $N$-way tensor ${\mathcal{X} \in \mathbb{R}^{I_1 \times I_2 \times \cdots \times I_N}}$ is a multidimensional array whose elements are indexed by $N$ indices. The number of free parameters in the tensor ${\mathcal{X}}$, $\prod_{n=1}^N I_n$, grows exponentially with $N$, a problem known as the {\em curse of dimensionality} (CoD). Any tensor $\mathcal{X}$ can be decomposed as a sum of $F$ rank-$1$ tensors as
\[\mathcal{X} = \sum_{f=1}^F{ \boldsymbol{\lambda}}(f) \mathbf{A}_1(:,f) \circ \mathbf{A}_2(:,f) \circ \cdots \circ \mathbf{A}_N(:,f),\]
where $\boldsymbol{\lambda} \in {\mathbb{R}}^{F}$, $\mathbf {A}_n \in {\mathbb{R}}^{I_{n}\times F}$, ${\bf A}_n(:,f)$ denotes the $f$-th column of matrix ${\bf A}_n$, and $\circ$ denotes the outer product (see Figure \ref{CPD}). We use the notation $\mathcal{X}= [\![\boldsymbol{\lambda}, \mathbf{A}_1,\ldots,\mathbf{A}_N]\!]$ to denote the decomposition.
A particular element of the tensor is given by
\[{\mathcal{X}(i_1,i_2,\ldots,i_N) = \sum_{f=1}^F \boldsymbol{\lambda}(f) \prod_{n=1}^N\mathbf{A}_n(i_n,f).}\]
The vectorized form of $\mathcal{X}$ can be expressed as
${\rm{vec}}(\mathcal{X})= (\odot_{n=1}^N \mathbf{A}_n)\boldsymbol{\lambda},$
where $\odot$ denotes the Khatri-Rao product and ${(\odot_{n=1}^N \mathbf{A}_n)= \mathbf{A}_N \odot \cdots \odot \mathbf{A}_1}.$
We can express the mode-$n$ matrix unfolding which is a concatenation of all mode-$n$ `fibers' of the tensor  as ${\mathcal{X}^{(n)} = (\odot_{k \neq n} \mathbf{A}_k){\rm{diag}}(\boldsymbol{\lambda}){\mathbf{A}_n}^T}$, where ${\rm{diag}}(\boldsymbol{\lambda})$ denotes the diagonal matrix with the elements of vector $\boldsymbol{\lambda}$ on its diagonal. 
When the number of rank-$1$ components is minimal, $F$ is called the rank of $\mathcal{X}$, and the decomposition is called {\em Canonical} Polyadic Decomposition (CPD), also known as PARAFAC or CANDECOMP~\cite{SiDeFu2017}. 
If the tensor can be well approximated by a low-rank CPD model, the CoD is alleviated as the number of free parameters drops to ${O}(NIF)$.
\subsection{Tensor Modeling of joint PMFs}
Let $Y$ be a random variable we wish to infer based on at most $N$ features $X_{V}= \{X_1,\ldots,X_N\}$, where $V := \left\{1,\ldots,N\right\}$, also called ground set. The joint Probability Mass Function (PMF) $P_{X_V,Y}$, of $X_V,Y$, can be represented by a probability tensor $\mathcal{X}$ where the size of each dimension is equal to the alphabet size $I_1,\ldots,I_{N+1}$ of the corresponding variable and the indexed elements represent the probability of the particular realization i.e., $\mathcal{X}(i_1,i_2,\ldots,i_{N+1} ) = {P}_{X_V,Y}(i_1,i_2,\ldots,i_{N+1})$. Every tensor $\mathcal{X}$ admits a CPD of finite rank and thus we can always express the joint PMF of $X_V, Y$ using a non-negative CPD model
\begin{equation}
  \mathcal{X}(i_1,\ldots, i_{N+1} ) 
 = \sum_{f=1}^F \boldsymbol{\lambda}(f) \prod_{n=1}^{N+1} \mathbf{A}_n(i_n,f),\label{eq:1}
\end{equation}
for high enough $F$~\cite{KaSiFu2018}. Equation~\eqref{eq:1} shows that every joint PMF admits a naive Bayes model \textit{interpretation} with bounded $F$.
\begin{figure}[!tbp]
  \centering
  \begin{minipage}[b]{0.22\textwidth}
  \centering
\includegraphics[width = 0.8 \columnwidth]{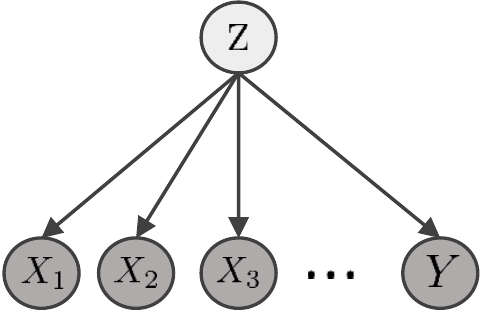}
\caption{Latent variable-naive Bayes model.}
\label{fig:NB}
  \end{minipage}
  \hfill
  \begin{minipage}[b]{0.25\textwidth}
\centering
\includegraphics[width = 0.7 \columnwidth]{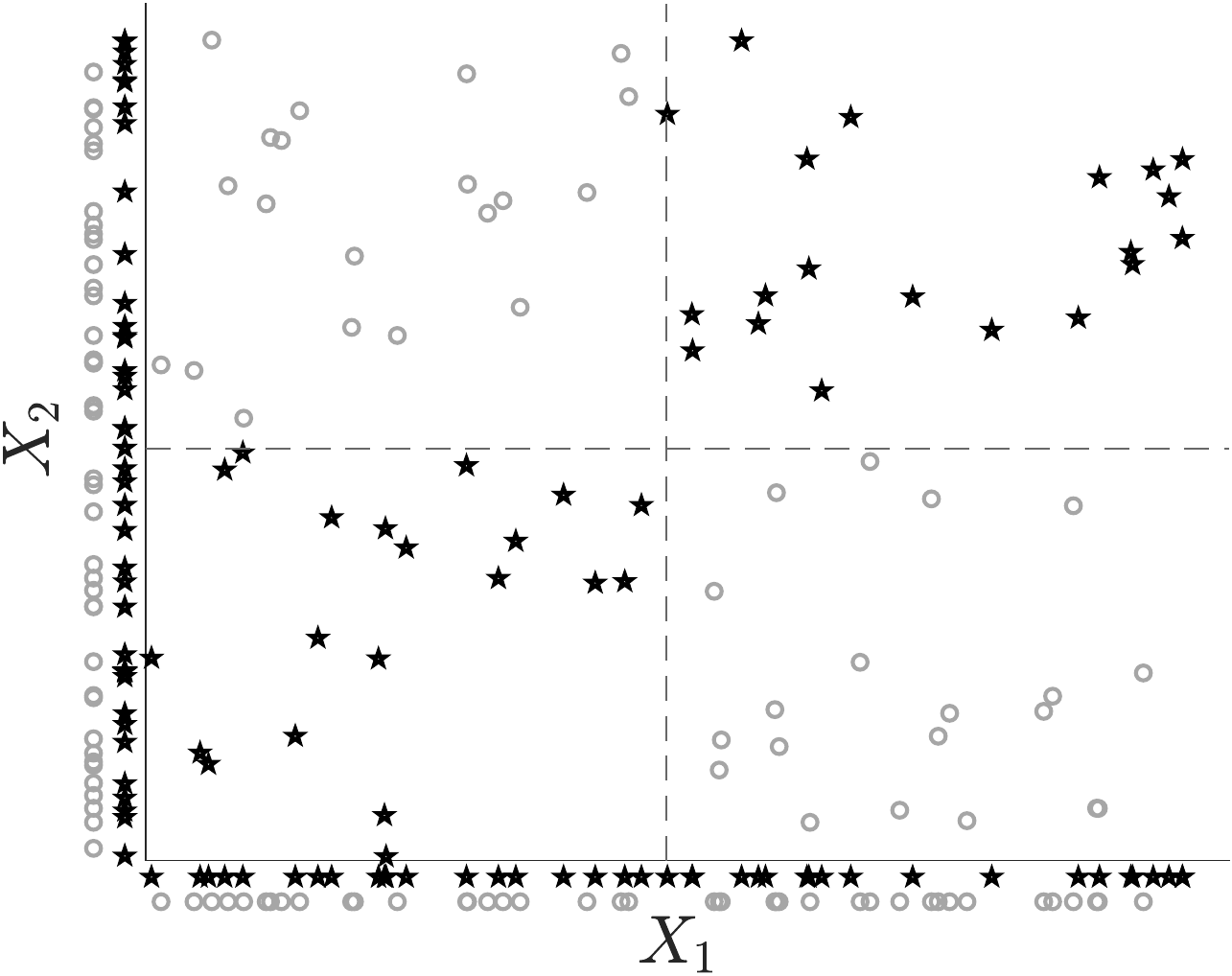}
\caption{$\protect X_1\perp \!\!\! \perp Y, X_2\perp \!\!\! \perp Y, \{X_1,X_2\} \not\!\perp\!\!\!\perp Y$.}
\label{inter}
  \end{minipage}
\end{figure}
The naive Bayes model assumes that there is a hidden random variable $Z$ taking $F$ values, such that given $Z=f$ the random variables $X_V,Y$ are conditionally independent i.e.,
\begin{align}
{P}_{X_V,Y}(i_1,&\ldots,i_{N+1})= \sum_{f=1}^F   P_Z(f){P}_{X_V,Y|Z}(i_1,\ldots,i_{N+1}|f) \nonumber\\
&= \sum_{f=1}^F P_Z(f)  {P}_{Y |Z} (i_{n+1} | f) \prod_{n=1}^{N} {P}_{X_n | Z} (i_n | f).\label{eq:2}
\end{align}
By variable matching between Equations \eqref{eq:1} and \eqref{eq:2} and upon defining 
$\mathbf{A}_n(i_n,f)
:= P_{X_n|Z}(i_n|f)$, for $n=1,\ldots,N$, $\mathbf{A}_{N+1}(i_{N+1},f):= P_{Y|Z}(i_{N+1} | f)$  and $\boldsymbol{\lambda}(f) := {P}_Z(f)$, we can  see that the naive Bayes model can be represented by a non-negative CPD model $\mathcal{X} = [\![\boldsymbol{\lambda},\mathbf{A}_{1},\ldots, \mathbf{A}_{N+1}] \!]$ 
with the constraints that matrices ${\bf A}_n$ are column-stochastic, and ${\mathbf{1}}^T \boldsymbol{\lambda} = 1$~\cite{KaSiFu2018, AmiKaSi2019}.
The observed data 
(so-called `manifest' variables) 
are generated through an unknown mapping expressed by the conditional distributions ${P}_{X_n|Z}, P_{Y|Z}$ and the prior distribution of the hidden variable ${P}_Z$. 
This model is also known as mixture of unigrams or latent class model and has been applied in many applications such as topic modeling~\cite{nigam}, clustering~\cite{Zhang2004} and crowdsourcing~\cite{ZhaCheZhou2014}.
In conclusion, {\em{any}} joint PMF can be represented by a latent variable model with just one hidden variable having $F$ possible states and therefore admits a non-negative CPD of bounded rank. Employing this model, we can alleviate the CoD by focusing on `principal components' of the joint distribution.


\subsection{Mutual Information and Submodularity}
The Shannon entropy of a random variable $X$ is defined as ${H(X)=-\sum_{x} P_X(x) \log P_X(x)}$ and it measures the amount of uncertainty in $X$. Given a second variable $Y$, we can quantify the uncertainty in $X$ after $Y$ has been observed using the conditional entropy ${H(X|Y)=-\sum_{x,y} P_{X,Y}(x,y) \log {P_{X|Y}(x|y)}}$. The Mutual Information (MI) between two random variables $X$ and $Y$ is defined as 
\begin{equation*}
I(X;Y)=\sum_{x,y} P_{X,Y}(x,y) \log \frac{P_{X,Y}(x,y)}{ {P_{X}(x)P_Y(y)}},
\end{equation*}
and measures how far the variables $X,Y$ are from being independent. Alternatively, we can view MI as $I(X;Y)=H(Y)-H(Y|X)$, which allows us to interpret MI as the reduction of the uncertainty about  $Y$ when we are provided with knowledge of $X$.
MI is symmetric in its arguments: $I(X;Y)=I(Y;X)=H(X)-H(X|Y)$.
Given the joint PMF $P_{X_V,Y}$ of ${X_V}= \{X_1,X_2,\ldots,X_N\}$ and $Y$, we consider the \textit{high-order mutual information} between a subset of features  $X_S$, $S \subseteq V$ and the variable $Y$ 
\begin{equation*}
f(S) =I(X_S;Y) =H(X_S)-H(X_S|Y),
\label{eq:homi}
\end{equation*}
which quantifies the expected reduction of uncertainty about $Y$ upon revelation of $X_S$. According to the non-decreasing property of MI, adding extra variables increases joint entropy, decreases conditional entropy, and increases information: $I(X_S;Y)\leq I(X_V;Y)$. MI~has been successfully employed in many feature selection methods due to the fact that it can capture complex relationships between the features and the target variable. However, selecting the optimal subset of features of cardinality $K$ that maximizes the high-order mutual information is known to be NP-hard~\cite{NP}. Additionally, the calculation of high-order mutual information requires a reliable estimate of the joint probability distribution. 

It has been shown that in the special case where the features are independent given the target variable $Y$ (which is a very restrictive and unrealistic assumption in practice), $f$ is monotone submodular~\cite{Krause,Krause2014SubmodularFM}. Submodular functions comprise a class of set functions $f:2^V\rightarrow \mathbb{R}$ that satisfy the diminishing returns property 
\begin{equation*}
f(A\cup\{x\})-f(A)\geq f(B\cup\{x\})-f(B)),
\end{equation*}
$ \forall A\subseteq B \subseteq V$ and  $x \in V \setminus  B$. This property states that adding an element to a smaller set results to larger increase in $f$ than adding it to a larger set. Moreover, if $f(A\cup\{x\})\geq f(A)$, $\forall A\subseteq V$ holds, the function is monotone submodular. \cite{Nemhauser1978} showed that the problem of maximizing a monotone submodular function $f$ subject to a cardinality constraint can be approximated with a constant factor $1-\frac{1}{\rm e}$ performance guarantee to the optimal solution of the NP-hard optimization problem using a simple greedy algorithm. Submodularity can be further exploited to accelerate the greedy implementation,  leading to an algorithm called lazy greedy with almost linear time complexity~\cite{lazy_greedy}.

\section{Problem Formulation}
Given a dataset $\mathcal{D}=\{\mathbf{x}_i,y_i\}_{i=1}^{M}$, of $M$ realizations of the random variables $X_V=\{X_1,X_2,\ldots,X_N\}$ (features) and the target variable $Y$ (label), we wish to infer a good subset $S\subseteq V$ of features within a budget, $|S|\leq K$ that best predicts $Y$. Ideally, $K$ is the {\em intrinsic} dimension of the dataset -- the minimum number of variables that carry sufficient information for accurately predicting the target variable $Y$. Intrinsic dimension can be alternatively viewed as the size of smallest feature-subset after which the MI between this subset and the target variable stops increasing. Initially, we formulate feature selection as an optimization problem by maximizing the MI between the features $X_S$ and the target variable $Y$, 
\begin{equation}
\aggregate{argmax}{S\subseteq V,|S|\leq K} f(S).\label{prob:initial}
\end{equation}
Instead of ranking each variable $X_n$ independently of the rest, this multivariate approach, which utilizes the high-order mutual information, evaluates features according to their joint information power, enabling us to detect redundant features. Feature interaction is significant in view of the fact that groups of several features acting simultaneously may be relevant, but not the individual features alone. In Figure~\ref{inter} features $X_1, X_2$ constitute an interaction group -- a set of features that appear to be irrelevant or weakly relevant with the class $Y$ individually, but if considered jointly, they correlate to the class. However, the number of candidate subsets is $\binom{N}{K}$, thus an exhaustive search is too costly and practically prohibitive even for a medium feature set size $K$.

Instead of solving optimization problem~(\ref{prob:initial}), we propose an intuitive and more efficient alternative approach. Since {\em every} joint distribution admits a latent variable - naive Bayes representation, via the CPD, we take an indirect path for determining the most informative features, through the latent variable $Z$. We propose using the mutual information as a metric to identify the subset of the `manifest' variables that can best identify the operational principal component of the distribution, or in other words, to best predict the latent variable $Z$ in the CPD model. 
The graphical model implies a dependence of the label $Y$ on the observed variable $X_n$ through the latent variable $Z$. 
Given $Z$, the features and the label $Y$ become conditionally independent, hence if we predict $Z$ from the features, predicting $Y$ from $Z$ is a simple task. 

In lieu of the initial $I(X_S;Y)$ maximization approach, we therefore propose solving the surrogate problem of selecting features by maximizing the MI between the selected features and the latent variable $Z$, i.e., 
\begin{align*}
&\aggregate{{argmax}}{S\subseteq V,|S|\leq K}g(S), \text{ where }\\ 
g(S) = I(X_S;Z)&= \sum_{x_S,z} P_{X_S,Z}(x_{S},z) \log \frac{ P_{X_S,Z}(x_{S},z)} {P_{X_S}( x_{S})P_Z(z)}.
\end{align*}
Employing a CPD model for the joint PMF $P_{X_V,Y}$, feature selection can be equivalently described as dropping out all but an optimal subset $S$ of $K$ edges tied to the $N$ features~(Fig.~\ref{fig:NB}). In terms of the CPD model, this means choosing a representative subset of factor matrices $\{\mathbf{A}_S\} \subset \{\mathbf{A}_V\}$ to form the reduced CPD model $\mathcal{X}'= [\![\boldsymbol{\lambda}, \{\mathbf{A}_S\}, \mathbf{A}_{N+1} ]\!]$.
\mybox{
\begin{Claim} 
\begin{equation*}
 g(S) - {\rm{const}}  \leq f(S) \leq g(S),
\end{equation*}
where ${\rm const}={I}(X_V;Z|Y)= {I}(X_V;\{Z,Y\})-{I}(X_V;Y)$.\end{Claim} 
\noindent \textbf{Proof:} Given $X_S,Y,\text{ and }Z$, the conditional mutual information is defined as $I(X_S;Z|Y)=I(X_S;\{Z,Y\})-I(X_S;Y)$. From the definition of the conditional MI and the latent variable model, for which it holds that $I(X_S;Y|Z)=0$, we get the following:
\begin{equation*}
\begin{aligned}
{I}(X_S;Y) &={I}(X_S; \{Y, Z\})-{I}(X_S;Z|Y) \\
  &={I}(X_S;Z)-{I}(X_S;Z|Y).
\end{aligned}
\end{equation*}
Since MI is always non-negative, we get that ${I}(X_S;Z|Y)\geq0$ and thus ${I}(X_S;Y)\leq{I}(X_S;Z)$, which means that our algorithm involves maximizing an upper bound of the direct approach. Furthermore, by the non-decreasing property of MI,  we  get that ${I}(X_S;Z|Y)~\leq{I}(X_V;Z|Y)$, and it holds that
\begin{equation*}
\begin{aligned}
{I}(X_S;Z)-{I}(X_V;Z|Y) & \leq{I}(X_S;Y)\leq{I}(X_S;Z)  \Leftrightarrow \\
{I}(X_S;Z)-{\rm const}& \leq{I}(X_S;Y)\leq{I}(X_S;Z) \Leftrightarrow \\
 g(S) - {\rm{const}} & \leq f(S) \leq g(S),
\end{aligned}
\end{equation*}
where ${\rm const}={I}(X_V;Z|Y)= {I}(X_V;\{Z,Y\})-{I}(X_V;Y)$.}
The double inequality shows that we are maximizing a surrogate function that is a constant band-gap away from the desired function. Intuitively, when the conditional entropy $H(Y|Z)$ is small, the band-gap is small. 

Like the original problem, the proposed alternative is NP-hard. The reason we propose it, however, is two-fold: first, given $Z$, all the $X$'s become irrelevant as far as $Y$ is concerned; and the above surrogate optimization problem where we aim to predict $Z$ involves the maximization of a monotonic submodular reward function subject to a cardinality constraint~\cite{Krause,Krause2014SubmodularFM}, which is not the case when our aim is to predict $Y$ directly from the regressors. Monotone submodular maximization subject to a cardinality constraint enjoys $1-\frac{1}{\rm e}$ approximation guarantee to the optimum solution, while simultaneously retaining extremely fast optimization~\cite{lazy_greedy}.

\section{Algorithm Description}
The proposed feature selection process, called Greedy Submodular Monotone optimization using CPD (GSM-CPD), consists of four steps, namely, PMF estimation of all variables, subset generation, MI evaluation and subset selection. 
\subsection{PMF Estimation}
In the first step, our algorithm utilizes a rank-$F$ approximation of the empirical joint PMF tensor $\widehat{\mathcal{X}}$, computed using Kullback-Leibler (KL) divergence as the fitting criterion. The empirical probability tensor, which is typically sparse, is formed by computing how often an event (a realization of the feature vector) occurred in the training set. The rank-$F$ approximation of the joint PMF captures the $F$ principal components of the distribution and is essential for the MI calculation process, which serves to evaluate the quality of the selected feature set $S$. Defining KL divergence between two probability tensors $\mathcal{X}$ and $\mathcal{Y}$ as 
\begin{equation*}
{\rm{D}_{KL}}( \mathcal{X} \Vert \mathcal{Y}):=\sum_{i_{1},\ldots,i_{N}}{ \mathcal{X}(i_{1},\ldots,i_{N} )}\log\frac{\mathcal{X}(i_{1},\ldots,i_{N} )}{\mathcal{Y}(i_{1},\ldots,i_{N})},
\end{equation*} 
we propose solving the following optimization problem 
\begin{align} 
{ \displaystyle \min_{\pmb{\lambda}, \mathbf{A}_{1},\ldots,\mathbf{A}_{N}}\displaystyle } & 
{ {\rm D_{KL}}{\Big( \widehat{\mathcal{X}} \Vert  [\![\pmb{\lambda}, \mathbf{A}_{\text{1}},\ldots, \mathbf{A}_{N} ]\!]\Big) } } \nonumber\\ { \text{subject to} } \ & { \quad \pmb{\lambda}\geq \mathbf{0}, {\mathbf{1}^{T}\pmb{\lambda} = 1, } }  \nonumber\\ &  { \quad \mathbf{A}_{n}\ge \mathbf{0},~ n=1\ldots N, } \nonumber\\ & { \quad \mathbf{1}^{T}\mathbf{A}_{n}= \mathbf{1}^{T}, ~ n = 1, \ldots, N}
\end{align} 
by employing the Expectation Maximization (EM) algorithm as described in \cite{ShaBhiSma2008} and~\cite{HuSi2017}. 
\begin{algorithm}[!t]
\caption{PMF Estimation}
\textbf{Input: } {Empirical PMF: $\widehat{\mathcal{X}}$, `Signal Rank' $F$ }\\
\textbf{Output:} {$ \mathcal{X} = [\![\boldsymbol{\lambda},
\mathbf{A}_1,\ldots,\mathbf{A}_N]\!]$}
\begin{algorithmic}[1]
\State Initial guess $ \mathcal{X} =[\![\boldsymbol{\lambda},
\mathbf{A}_1,\ldots,\mathbf{A}_N]\!]$
\State  $\widehat{\mathcal{Y}}\leftarrow {\widehat{\mathcal{X}}} /$  $ [\![\boldsymbol{\lambda},
\mathbf{A}_1,\ldots,\mathbf{A}_N]\!]$
\While{termination condition not met}
\ForAll{$f$}
\State $ \boldsymbol{\lambda}(f) \leftarrow \boldsymbol{\lambda}(f) \widehat{\mathcal{Y}}  \times_{1}{\mathbf{A}_1(:,f)} \cdots \times_{N}{\mathbf{A}_N(:,f)} $
\EndFor
\State
\textbf{for all} $n$ \textbf{update in parallel}
\State \quad~ $\mathbf{A}_n \leftarrow  \mathbf{A}_n * {\rm MTTKRP}(\widehat{\mathcal{Y}}, \{\mathbf{A}_n\}_{n=1}^N,n)$
\State 
\textbf{end for}
\State $ \widehat{\mathcal{Y}} \leftarrow \widehat{\mathcal{X}}/$  $ [\![\boldsymbol{\lambda},
\mathbf{A}_1,\ldots,\mathbf{A}_N]\!] $
\EndWhile
\State $\mathcal{X} \leftarrow [\![\boldsymbol{\lambda},
\mathbf{A}_1,\ldots,\mathbf{A}_N]\!] $\;
\end{algorithmic}
\label{alg:GSM-CPD}
\end{algorithm}
At each iteration, EM updates the factors simultaneously, making the algorithm easily parallelizable. The exact updates for $\boldsymbol{\lambda}, \mathbf{A}_n$ are shown in Algorithm~\ref{alg:GSM-CPD}. Notation~$\times_{n}$ stands for the $n$-mode product of a tensor with a matrix, MTTKRP denotes the $n$-mode matricized tensor times Khatri-Rao product, and $/$ stands for element-wise division. Note that the complexity of this operation is $\mathcal{O}(M)$. Here, $M$ denotes the number of samples which is approximately equal to the non-zero elements of tensor $\widehat{\mathcal{X}}$ .

Determining the rank $F$ of tensor ${\mathcal{X}}$ is an NP-hard problem~\cite{hillar2013}. Essentially, instead of detecting the exact rank, we are interested in fitting a model that has `meaningful' number of components -- the useful `signal rank', which is determined by cross-validation techniques. The per iteration complexity of the algorithm is dominated by the $\boldsymbol{\lambda}$-update, which is of  $\mathcal{O}(M {F})$ complexity,  and by each $\mathbf{A}_n$-update, which is also  $\mathcal{O}(M{F})$ complexity.
\subsection{Incremental Greedy Feature Selection}
After fitting a low-rank CPD model to the empirical PMF, we employ a forward greedy algorithm~(Alg.~\ref{alg:greedy}) for the problem of subset selection, i.e, $\displaystyle \max_{{S\subseteq V,|S|\leq K}} {g(S).} 
$
During the subset generation procedure, candidate feature subsets are generated for evaluation based on the MI. Starting with an empty set $S=\emptyset$, the algorithm incrementally builds a solution. At iteration $i$, it selects the feature $s_i$ that improves the current solution the most, according to information gain 
\begin{equation*}
s_i = \aggregate{{argmax}}{s\in V\setminus S} {I(X_{S \cup \{s\}};Z)-I(X_S;Z)},
\end{equation*} 
and adds it to the current set $S \leftarrow S\cup \{s_i\}$. The process of subset generation and evaluation is repeated until $|S|= K$. The output of the greedy algorithm is always a set $S$ such that 
\begin{equation*}
I(X_S;Z)\geq \left ( 1-\frac{1}{e} \right ) I(X_{S^{\star}}; Z ), 
\end{equation*}
where $S^{\star}$ is the optimal solution i.e., the set maximizing $g(S)$ among all size-$K$ sets.
\begin{algorithm}[!t]
\caption{Incremental Greedy Feature Selection}
\label{alg:greedy}
\textbf{Input: } {$K$: Number of features; $\mathcal{X}$: Joint PMF tensor}\\
\textbf{Output: } $S$: Estimated subset of features
\begin{algorithmic}[1]
\State $V = \{1,2,\ldots,N \}$
\State $ S=\emptyset$
\While{$|S|< K$}
\For{\textbf{all} $s \in V \setminus S$}
\State $ MI(s) = {I}(X_{S\cup s};Z)$
\EndFor
 \State $s_i \leftarrow \text{feature with maximum}~MI$
\State $S \leftarrow \{S \cup s_i \}$\;
\EndWhile
\end{algorithmic}
\end{algorithm}
\begin{figure*}
\centering
\subfloat{{\includegraphics[width=0.25\textwidth]{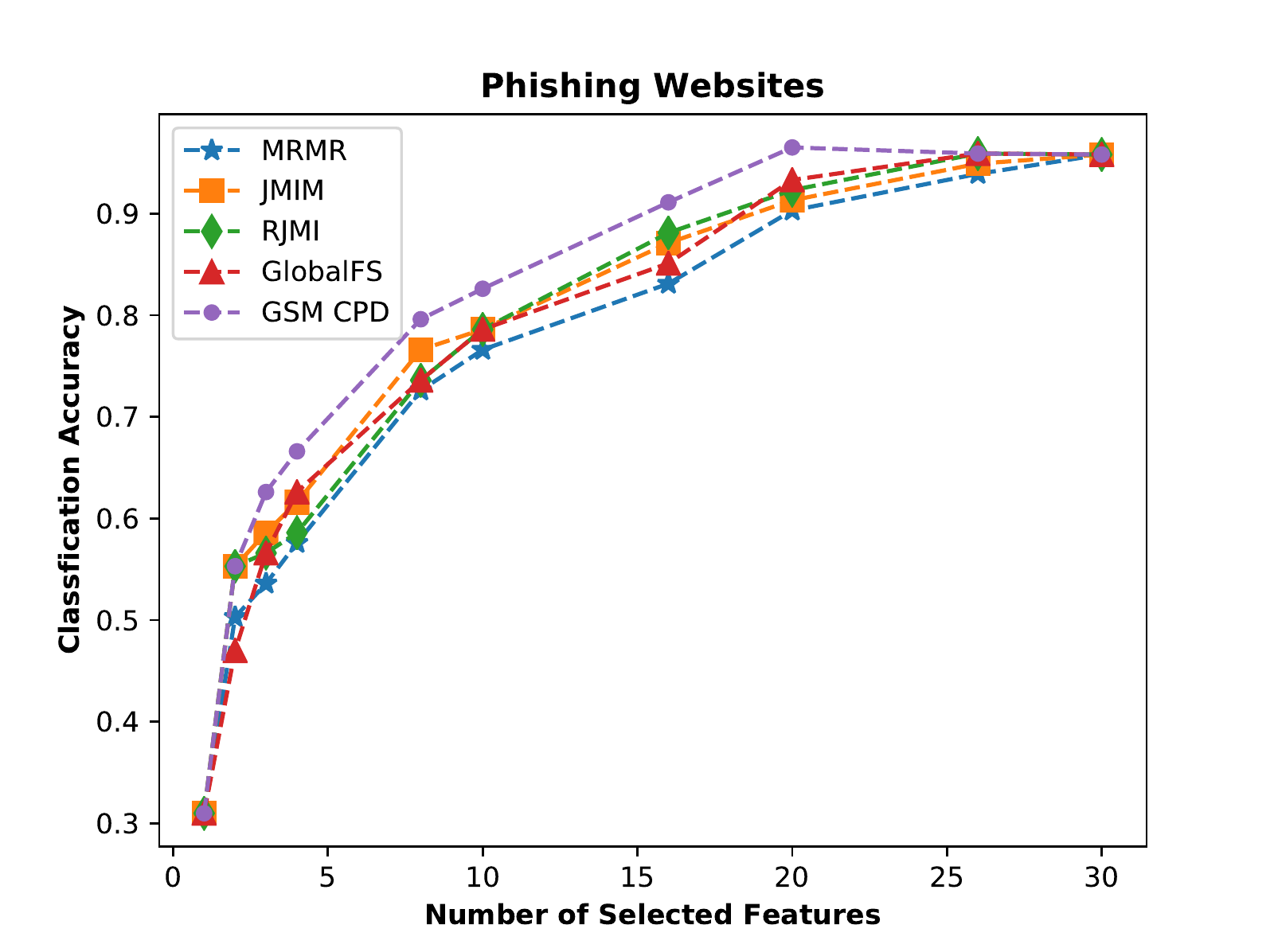}\label{fig1}}}\hfill
 \subfloat{{\includegraphics[width=0.25\textwidth]{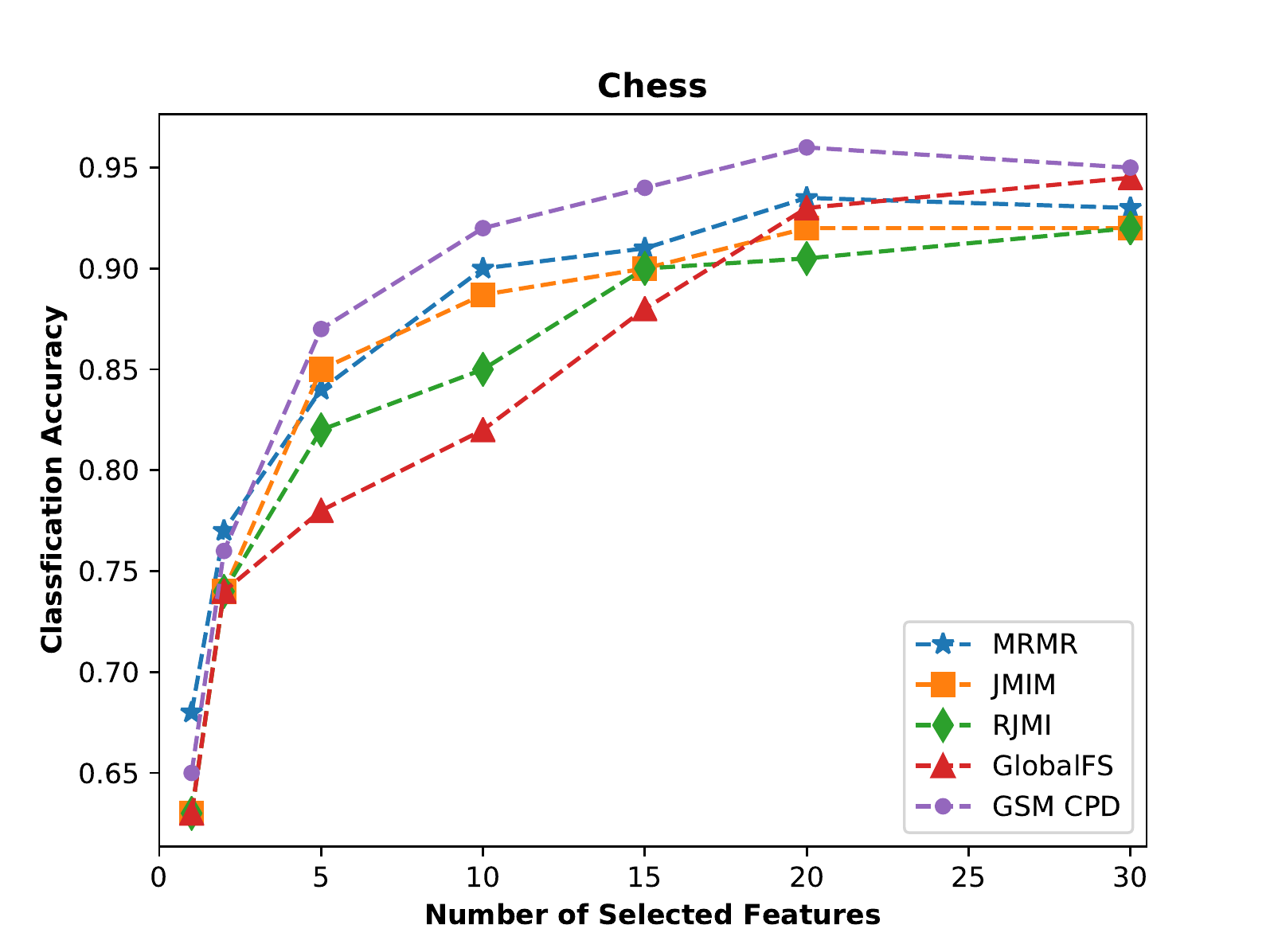}\label{fig2}}}\hfill
\subfloat{{\includegraphics[width=0.25\textwidth]{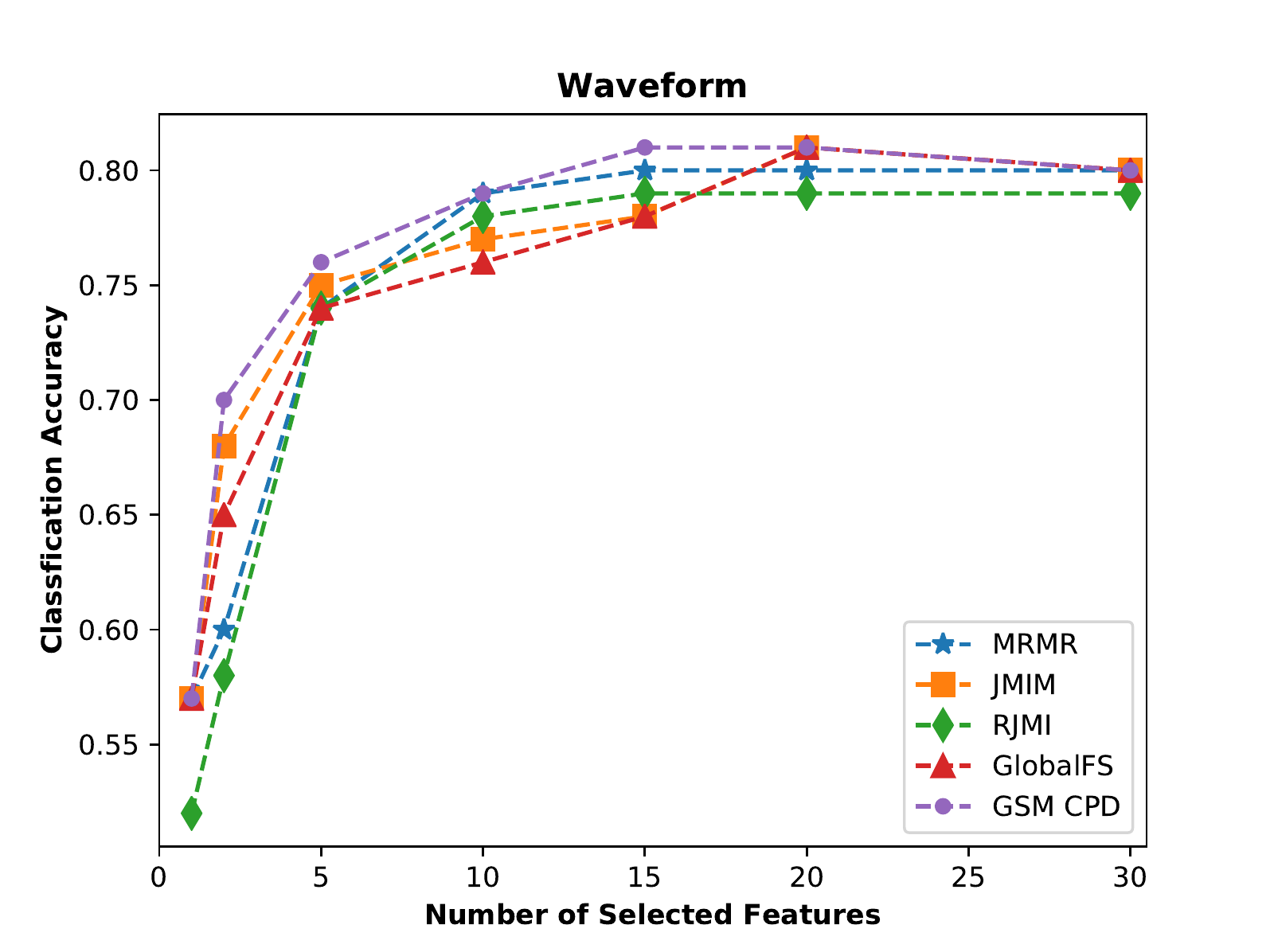}\label{fig3}}}\hfill
  \subfloat{{\includegraphics[width=0.25\textwidth]{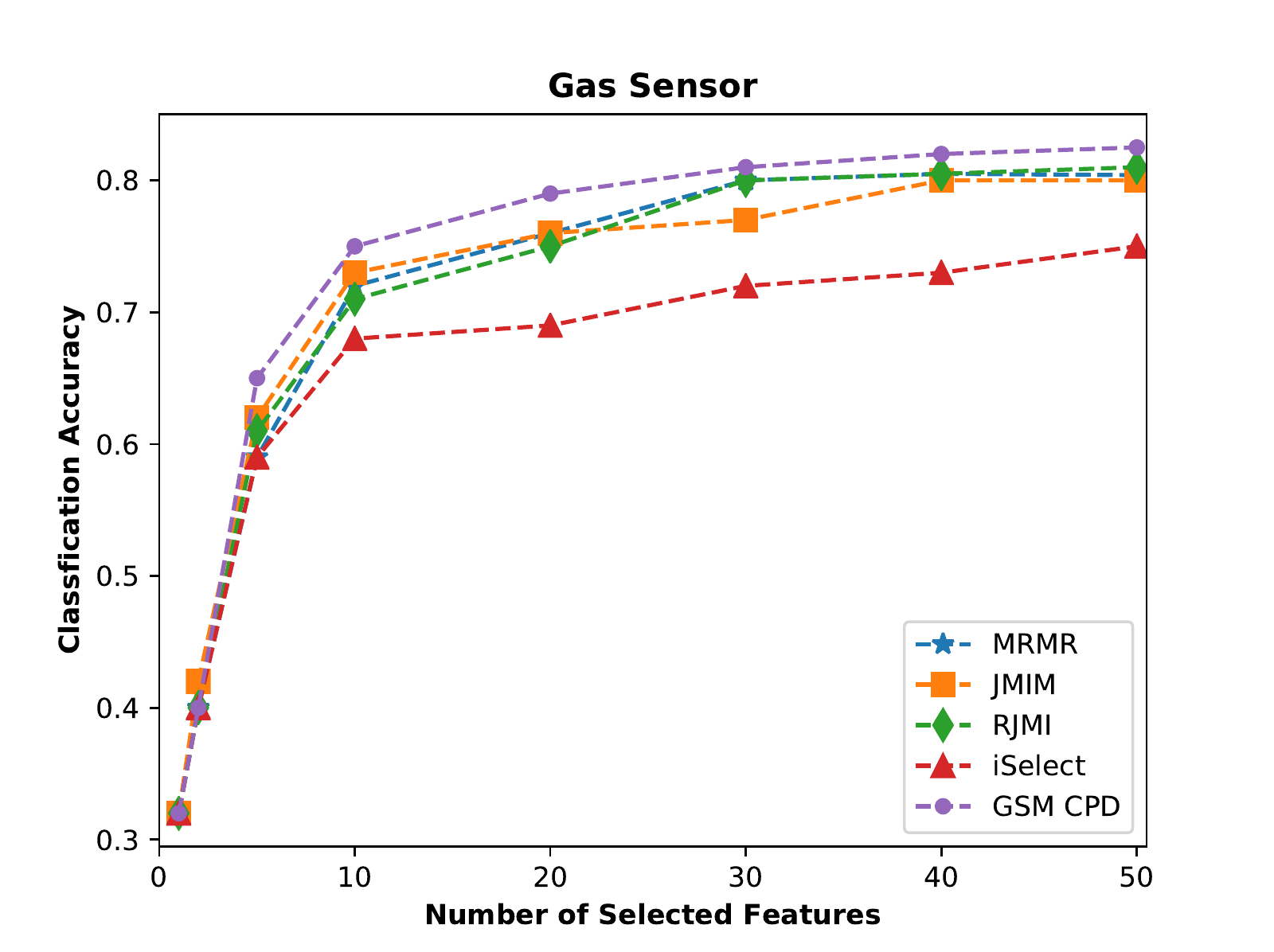}\label{fig4}}}
\\ \vspace{-0.15in} 
 \subfloat{{\includegraphics[width=0.25\textwidth]{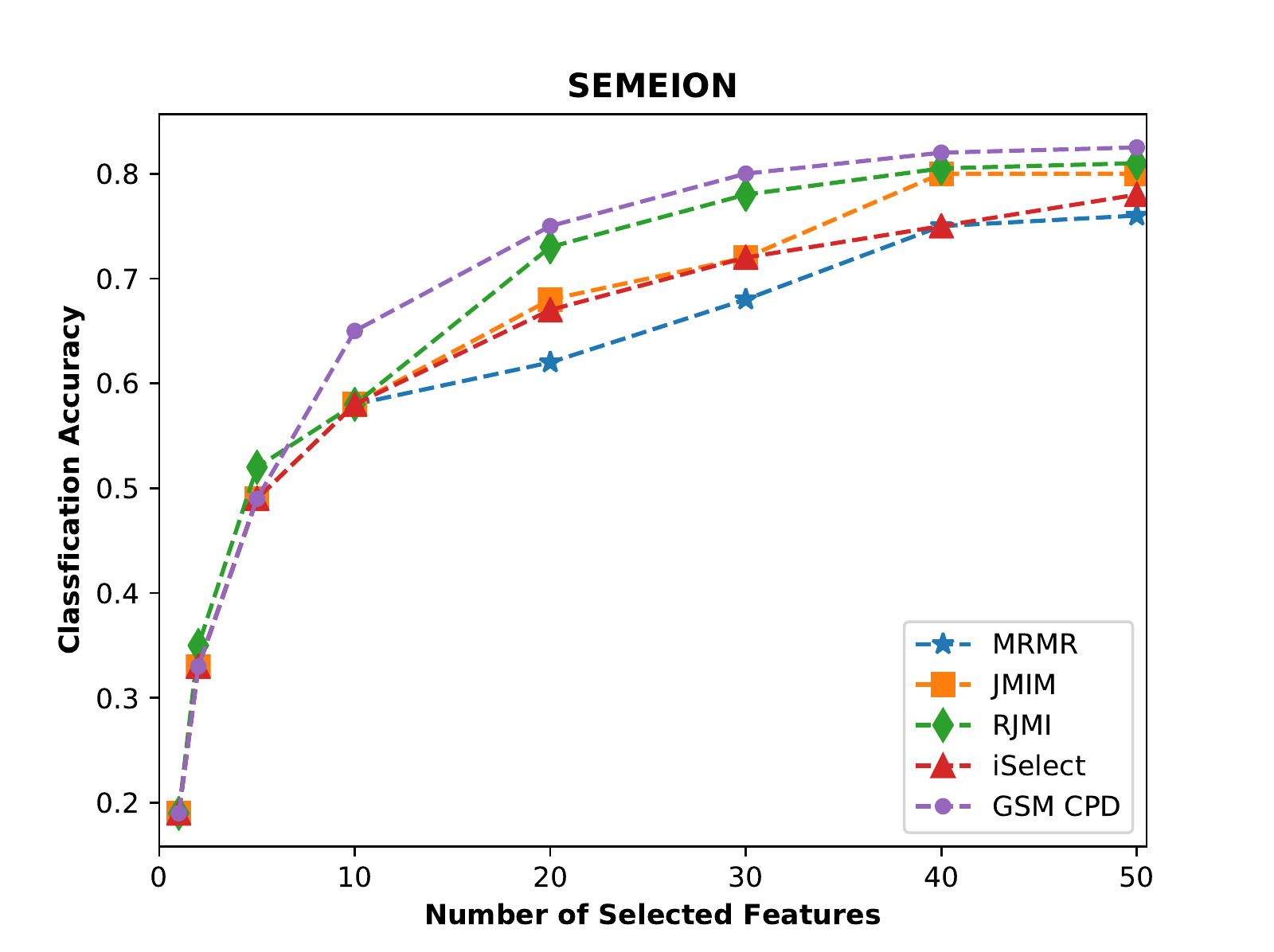}\label{fig5}}}
    \subfloat{{\includegraphics[width=0.25\textwidth]{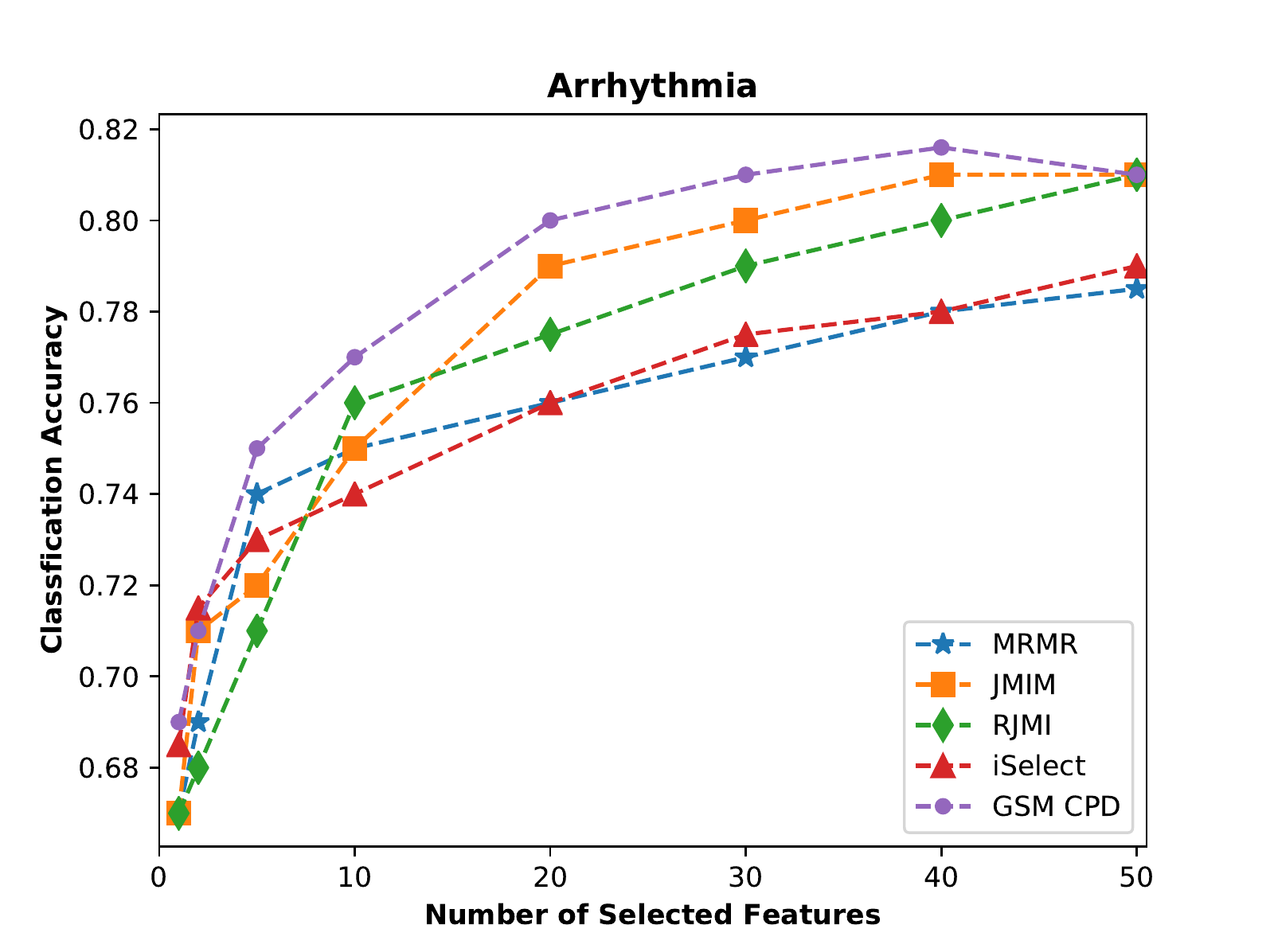}\label{fig6}}}
    \subfloat{{\includegraphics[width=0.25\textwidth]{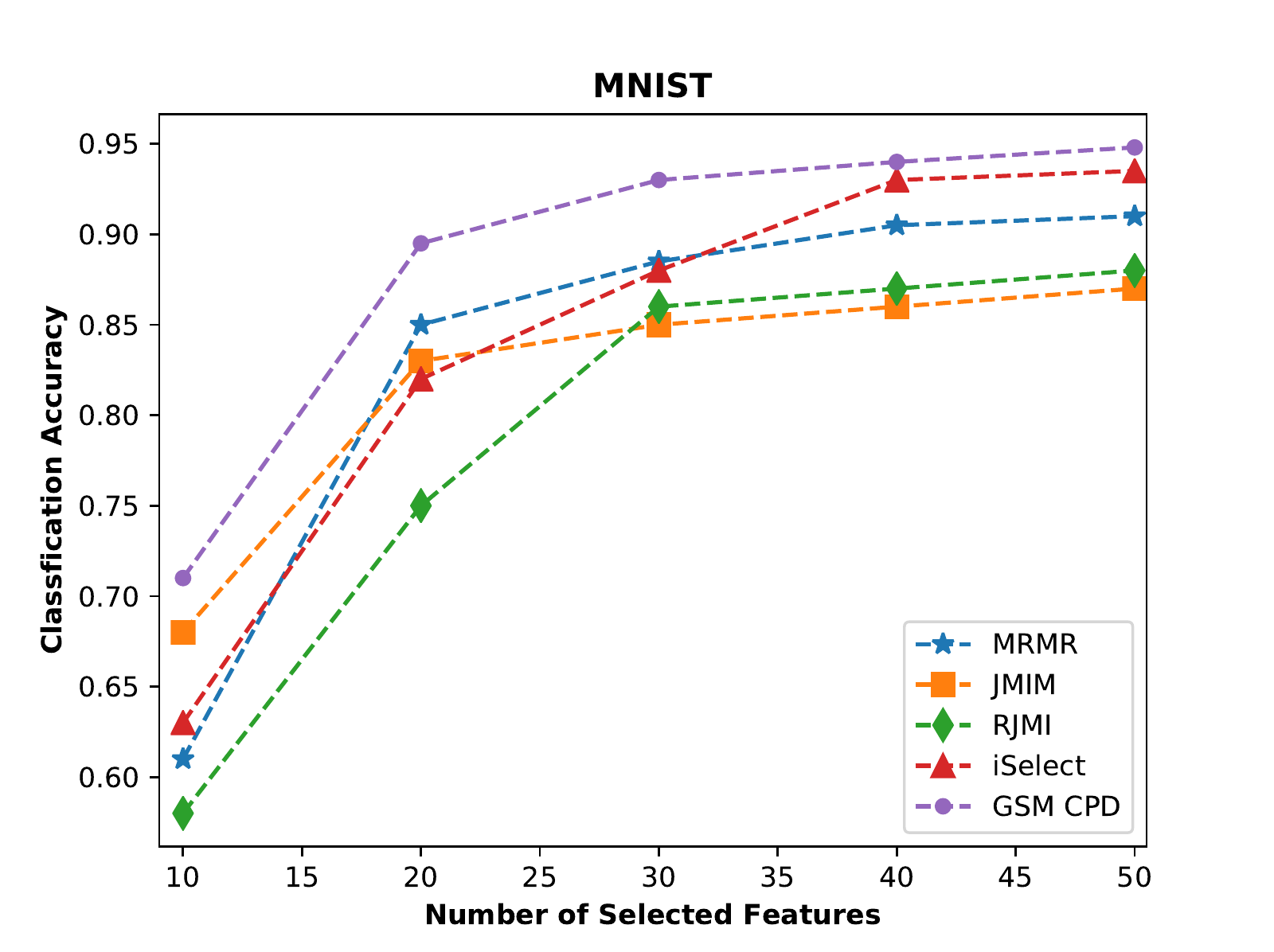}\label{fig7}}}
    \subfloat{{\includegraphics[width=0.25\textwidth]{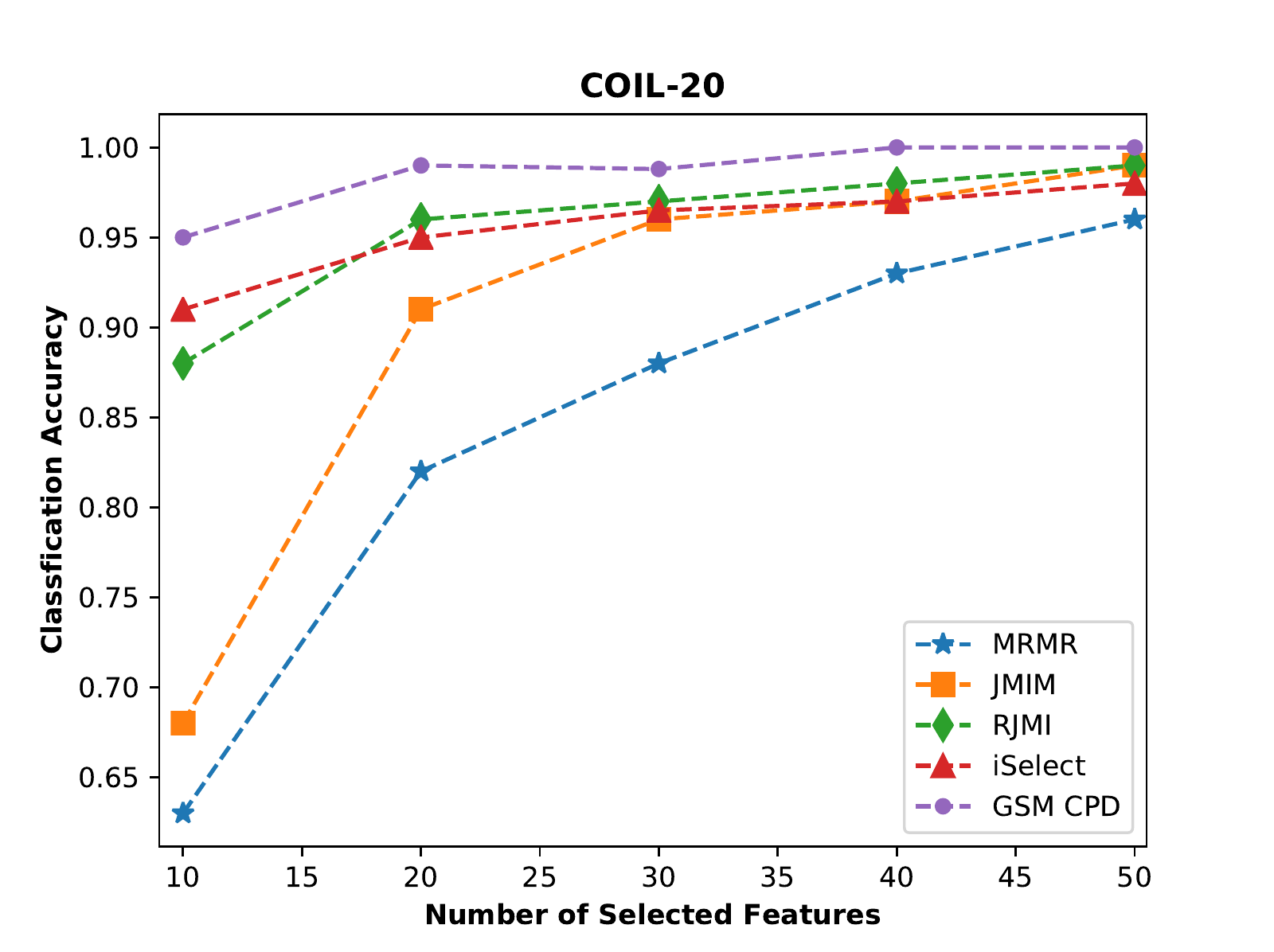}\label{fig8}}}
    \caption{1-NN classification  accuracy  versus  the  number  of selected  features.}
\label{GSM_compare}
\end{figure*}
Without loss of generality, assume that at iteration~$k \leq K$, $X_S=\{X_1, X_2,\ldots,X_k\}$. We can evaluate the MI between $X_S$ and the latent variable $Z$ as 
\begin{equation*}
\begin{aligned}
{I}&(X_S ;Z) = \sum_{x_S,z} {P}_{X_S,Z}(x_S,z)  \log \frac{ {P}_{X_S,Z}(x_S,z)}{{P}_{X_S}(x_S)P_Z(z)} \\
&=  \mathbf{1}^T\left[  (\odot_{n=1}^k \mathbf{A}_n)  {\rm diag}({\boldsymbol \lambda}) * \log{\frac{  (\odot_{n=1}^k \mathbf{A}_n)  {\rm diag} ({\boldsymbol \lambda})}
{[(\odot_{n=1}^k \mathbf{A}_n) \boldsymbol{\lambda}] \circ \boldsymbol{\lambda}}}\right]\mathbf{1}, 
\end{aligned}
\end{equation*}
where $*$ indicates the matrix Hadamard product and the logarithm is computed element-wise. In case of large $k$ the above computation is prohibitive. It can be simplified using the fact that under the naive Bayes model, MI is given by
\begin{equation*}
\begin{aligned}
{I}(X_S;Z) &= H(X_S) - H(X_S | Z) = H(X_S) - \sum_{n=1}^K H(X_n | Z).
\end{aligned}
\end{equation*}
In terms of the CPD factors, the joint entropy of $X_S$ is given by
\begin{equation*}
\begin{aligned}
H(X_S) &=-\sum_{x_S}{P}_{X_S}(x_S)\log{P}_{X_S}(x_S) \\ 
&=- \sum_{i_1,\ldots,i_K}{\mathcal{X}}(i_1,\ldots,i_k)\log  \mathcal{X}(i_1,\ldots,i_k) \\
&=- \mathbf{1}^T \left[ (\odot_{n=1}^K \mathbf{A}_n) \boldsymbol{\lambda} * \log (\odot_{n=1}^K \mathbf{A}_n) \boldsymbol{\lambda} \right]  \mathbf{1},
\end{aligned}
\end{equation*}
and the conditional entropy for each variable $X_n$ is given by
\begin{equation*}
\begin{aligned}
H(X_n | Z) &=-\sum_{x_n,z}{P}_{X_n,Z}(x_n,z)\log{P}_{X_n|Z}(x_n|z) \\
&=-\sum_{i_n,f} \mathbf{A}_n(i_n,f)\boldsymbol{\lambda}(f) \log  \mathbf{A}_n(i_n,f) \\
&=- \mathbf{1}^T \left[( \mathbf{A}_n{\rm diag}(\boldsymbol{\lambda}) )* \log  \mathbf{A}_n \right] \mathbf{1}.
\end{aligned}
\end{equation*}
Calculating the MI function can be expensive due to the computational bottleneck of the joint entropy, which requires $I^k$ evaluations. To overcome this issue, we can take advantage of the fact that $H(X_S)=- \mathbb{E} [\log{P}_{X_S} (X_S)]$ to calculate an approximation for this term, by drawing samples from the joint distribution. We randomly sample $T$ values of the latent variable $Z$ according to its distribution $\boldsymbol{\lambda} $ and given each value $f$ we similarly sample from the $f$-th column of each factor matrix $\mathbf{A}_n(:,f)$, $\forall n \in S$. We calculate the probability of this particular realization via the probability tensor $ \mathcal{X}'= [\![\boldsymbol{\lambda}, \{\mathbf{A}_S\}]\!]$. After transforming to logarithmic scale, we sum up the observations and we normalize by $T$ to get the expected value. Note that, in our experiments drawing $1000-5000$ samples is sufficient for a well approximated joint entropy. The per iteration complexity of the algorithm is determined by the calculation of the joint entropy which is of $\mathcal{O}(NFT)$ complexity and the conditional entropy computation which is of $\mathcal{O}(NFI)$ complexity. In total, the complexity of the algorithm is $\mathcal{O}(KNF(T+I))$.


\section{Experimental Study}

\noindent {\bf Results and Discussion:} We conducted experiments on real-world datasets to assess the performance of the proposed GSM-CPD sequential forward feature selection framework against various supervised information-theoretic based feature selection algorithms, that are representative of the state-of-art. See \cite{LiCheWa2018} for a recent tutorial overview. All datasets are from the UCI machine learning repository~\cite{UCI}. A summary of the selected
datasets is presented on Table \ref{tab:table2}. For each dataset, continuous features are  discretized using an equal-width strategy into $5$ bins, while already discrete features, or features with a categorical range are left untouched.
\begin{table}[H]
  \begin{center}
    \begin{tabular}{|c | c | c |c|}
    \hline
 \textbf{Datasets} & \textbf{N} & \textbf{M}& \textbf{C}\\ 
      \hline
    Phishing Websites & 30 & 2456&2\\ 
    Chess (King-Rook vs. King-Pawn) & 36 & 3196& 2\\ 
Waveform (Version 2)&40 & 5000&3\\ 
 Gas Sensor Array Drift &128&13910&3\\ 
 Semeion &256&1593&10\\
 Arrhythmia&279&452&16\\
 MNIST &784&80000&10\\
 COIL-20 &1024&1440&20\\
  \hline
    \end{tabular}
\medskip
 \caption{Summary of bench-mark datasets.}
\label{tab:table2}
  \end{center}
\end{table}
\noindent {\bf Experimental Settings:} Numerous methods have previously been proposed for feature selection. Most MI-based techniques are greedy methods that make use of low dimensional MI quantities due to the difficulty associated with estimating the high dimensional distributions from limited samples. To address this problem recent techniques consider interactions among more than two variables, by estimating/ approximating higher-dimensional mutual information quantities.
We selected $4$ state-of-the-art information-theoretic feature selection methods (MRMR~\cite{mrmr}, JMIM~\cite{bennasar2015feature}, RJMI~\cite{yu2019multivariate}, and GlobalFS/iSelect~\cite{vinh2014reconsidering}), and compared them to our method. \begin{enumerate}
\item MRMR: The Maximum Relevance Minimum Redundancy approach is an information-theoretic based method, where $S$ grows sequentially--one feature is added at a time based on its MI with $Y$ while minimizing the dependency among the features already selected. MRMR is a greedy algorithm like our method, but with a coarser criterion~\cite{mrmr}.
\item JMIM: The Joint Mutual Information Maximization approach employs both the maximum of the minimum’ approximation, which is integrated approximation of the relevancy and redundancy, and the joint mutual information between candidate features, selected features and the class, to addresses the problem of overestimation the significance of some features~\cite{bennasar2015feature}.
\item Rényi-based JMI (Rényi’s $\alpha$-order based joint MI Maximization): Instead of building upon classic discrete Shannon’s information quantities, authors in \cite{yu2019multivariate} define a multivariate extension of the matrix-based Rényi’s $\alpha$-order joint entropy, the method allows estimating the multivariate MI with repsect to  a desired variable $Y$, without  evaluating  the underlying PMF.
\item  GlobalFS/iSelect: Authors in ~\cite{vinh2014reconsidering} aim to find a set of features that jointly maximizes the mutual information with the class variable by subtracting from the plugin estimator a corrective term based on a $\chi^2$ statistical test.
\end{enumerate}

The proposed approach, GSM-CPD, is implemented in MATLAB using the Tensor Toolbox~\cite{TTB_Sparse} for tensor operations. For each experiment we split the dataset such that $70\%$ of the data samples is used for training and $30\%$ for testing, and run $10$ Monte-Carlo simulations. An appropriate rank $F$ for our model is found using $5-$fold cross-validation. For each dataset, we fit CPD models of different ranks, $F \in \{5,10,15,20,30\}$, and choose the one which on average minimizes the misclassification error on the validation set. 
After extracting the optimal subset $S$ of features using each method, each subset of features is evaluated in terms of the classification performance of the $1-$nearest-neighbor ($1-$NN) classifier as a conventional way of evaluating supervised feature selection methods. We report the mean classification accuracy of each feature selection method in various numbers of selected features, $K=1,\ldots, 50$. {For the four largest datasets, the feature selection process is realized with the following modification. At iteration~$k+1, k \leq K$, where the current subset of selected features is $X_S=\{X_1, X_2,\ldots,X_k\}$, we remodel the PDF of the subsets $X_{S \cup \{s\}}, s\in V\setminus S$ of variables and select the one that maximizes the MI.}

{Figure~\ref{GSM_compare} depicts the predictive performance using the above feature selection methods followed by the $1-$NN classifier as a function of the number of selected features. For each dataset, beginning with only a few features, the selected feature set is gradually grown until reaching $30$ or $50$ features, depending on the dataset dimensionality. The results demonstrate the superior performance of GSM-CPD as a feature selection strategy. In almost all of the datasets, and especially for MNIST and COIL$-20$, GSM-CPD has a clear lead compared to the baselines which demonstrates the capability of our algorithm to select the most informative features. It is important to note that, using the proposed method, for datasets Phishing Websites, Chess, Waveform, and Arrhythmia, better performance is achieved utilizing a smaller subset than the maximum $K$ considered. Therefore, we can design a classifier based on the GSM-CPD selected features without sacrificing accuracy. For the rest of the datasets (Gas Sensor, SEMEION, MNIST, and COIL$-20$), the best performance is achieved when using all $K$ features. Even then, there is always another reduced-dimension close in performance, which uses a smaller fraction of the original features. It is also notable that by tuning the value of $K$ we can identify the intrinsic dimension of the dataset. For example, for Gas Sensor the minimum number of features needed for an acceptable performance is $20$. Selecting less than $20$ features yields a significant degradation in prediction performance. In all cases, our GSM-CPD feature selection method appears to be very effective in feature selection, and often close to optimal in terms of classification accuracy.}
\section{Conclusions}
In this paper, we presented a novel low-complexity approach for identifying the most predictive subset of variables without compromising  classification accuracy. In the first step, we model the joint PMF of the complete set of variables using a latent variable model following the naive Bayes hypothesis. 
In our present context it naturally suggests a monotone submodular surrogate optimization problem that is amenable to greedy optimization with performance guarantees. This gives rise to the proposed GSM-CPD feature selection approach.  Experiments on real-world data show that GSM-CPD can outperform well-appreciated baseline methods by a significant margin. 

\balance{}
\bibliography{references}{}
\bibliographystyle{IEEEtran}

\end{document}